%% file: main.tex
\documentclass[conference]{IEEEtran}
\IEEEoverridecommandlockouts
\usepackage{cite}
\usepackage{amsmath,amssymb,amsfonts}
\usepackage{algorithmic}
\usepackage{graphicx}
\usepackage{textcomp}
\usepackage{tabularx}
\usepackage{xcolor}
\def\BibTeX{{\rm B\kern-.05em{\sc i\kern-.025em b}\kern-.08em
    T\kern-.1667em\lower.7ex\hbox{E}\kern-.125emX}}

\input{include}
\input{acronym}
\usepackage{url}
\makeatletter
\def\ps@IEEEtitlepagestyle{%
  \def\@oddfoot{\mycopyrightnotice}%
  \def\@evenfoot{}%
  \def\@oddhead{}
  \def\@evenhead{}%
}
\makeatother
\def\mycopyrightnotice{%
  \begin{minipage}{\textwidth}
  \centering \scriptsize
  \copyright 2026 IEEE.  Personal use of this material is permitted.  Permission from IEEE must be obtained for all other uses, in any current or future media, including reprinting/republishing this material for advertising or promotional purposes, creating new collective works, for resale or redistribution to servers or lists, or reuse of any copyrighted component of this work in other works.
  \end{minipage}
}
\makeatother
\begin{document}

\title{
PanLUNA: An Efficient and Robust Query-Unified Multimodal Model for Edge Biosignal Intelligence}

\author{\IEEEauthorblockN{
Marija Zelic\IEEEauthorrefmark{1},
Anna Tegon\IEEEauthorrefmark{1},
Yawei Li\IEEEauthorrefmark{1},
Thorir Mar Ingolfsson\IEEEauthorrefmark{1},
Luca Benini\IEEEauthorrefmark{1}\IEEEauthorrefmark{2}}
\IEEEauthorblockA{\IEEEauthorrefmark{1}Integrated Systems Laboratory, ETH Z{\"u}rich, Z{\"u}rich, Switzerland}
\IEEEauthorblockA{\IEEEauthorrefmark{2}DEI, University of Bologna, Bologna, Italy}
\thanks{Corresponding email: \{ategon,thoriri\}@iis.ee.ethz.ch}
}
\maketitle

\begin{abstract}
Physiological foundation models (FMs) have shown promise for biosignal representation learning, yet most remain confined to a single modality such as EEG, ECG, or PPG, largely because paired multimodal datasets are scarce. In this paper, we present PanLUNA, a compact 5.4M-parameter pan-modal FM that jointly processes EEG, ECG, and PPG within a single shared encoder. Extending LUNA's channel-unification module, PanLUNA treats multimodal channels as entries in a unified query set augmented with sensor-type embeddings, enabling efficient cross-modal early fusion while remaining inherently robust to missing modalities at inference time. Despite its small footprint, PanLUNA matches or exceeds models up to 57$\times$ larger: 81.21\% balanced accuracy on TUAB abnormal EEG detection and state-of-the-art 0.7416 balanced accuracy on HMC multimodal sleep staging. Quantization-aware training with INT8 weights recovers $\geq$96\% of full-precision performance, and deployment on the GAP9 ultra-low-power RISC-V microcontroller for wearables achieves 325.6\,ms latency and 18.8\,mJ per 10-second, 12-lead ECG inference, and 1.206\,s latency at 68.65\,mJ for multimodal 5-channel sleep staging over 30-second epochs.
\end{abstract}

\input{Sections/Introduction}
\input{Sections/Methods}

\input{Sections/Results}

\section{Conclusion}\label{ch:conclusion}
We presented PanLUNA, a 5.4M-parameter multimodal physiological foundation model that unifies EEG, ECG, and PPG processing within a single encoder by extending LUNA's channel-unification mechanism to cross-modal fusion via sensor-type embeddings. PanLUNA matches or exceeds models up to 57$\times$ larger on established EEG and ECG benchmarks and achieves state-of-the-art multimodal sleep staging on HMC, while remaining inherently robust to missing modalities at inference time. Quantization-aware training enables aggressive compression down to INT2 weights with graceful degradation, and deployment on the GAP9 ultra-low-power MCU achieves 325.6\,ms latency at 18.8\,mJ per inference for unimodal ECG, and 1.206\,s at 68.65\,mJ for multimodal EEG+ECG sleep staging, demonstrating, for the first time, that a multimodal physiological FM can operate within the power envelope of a wearable device across both unimodal and multimodal configurations. Future work will explore scaling to additional modalities and clinical validation in continuous monitoring scenarios.

\section*{Acknowledgment}
This work was supported by a grant from the Swiss National Supercomputing Centre (CSCS) under project IDs lp12 and lp160 on Alps. We acknowledge the CINECA award under the ISCRA initiative (project IsB32) for the availability of HPC resources.

\clearpage
\bibliographystyle{IEEEtranCustom}
\bibliography{bib}

\end{document}

%% file: include.tex
\usepackage{comment}

\usepackage{nohyperref}

\definecolor{matplotlib0}{HTML}{1f77b4}
\definecolor{matplotlib1}{HTML}{d62728}
\definecolor{matplotlib2}{HTML}{2ca02c}
\definecolor{matplotlib3}{HTML}{ff7f0e}
\definecolor{matplotlib4}{HTML}{9467bd}
\definecolor{matplotlib5}{HTML}{8c564b}
\definecolor{matplotlib6}{HTML}{e377c2}
\definecolor{matplotlib7}{HTML}{7f7f7f}
\definecolor{matplotlib8}{HTML}{bcbd22}
\definecolor{matplotlib9}{HTML}{17becf}

\usepackage{mathtools} 

\usepackage{subcaption}

\usepackage{booktabs}
\usepackage{array}
\usepackage{multirow}
\usepackage{colortbl}
\usepackage{tablefootnote}
\usepackage{threeparttable}
\usepackage{makecell}

\usepackage[acronym, style=super, nonumberlist]{glossaries}

\usepackage{pgfplots}
\usepgfplotslibrary{groupplots}
\definecolor{color0}{rgb}{0.12156862745098,0.466666666666667,0.705882352941177} 
\definecolor{color1}{rgb}{1,0.498039215686275,0.0549019607843137}
\definecolor{color2}{rgb}{0.172549019607843,0.627450980392157,0.172549019607843} 
\definecolor{color3}{rgb}{0.83921568627451,0.152941176470588,0.156862745098039} 
\definecolor{color4}{rgb}{0.580392156862745,0.403921568627451,0.741176470588235}
\definecolor{colorblue}{rgb}{0.12156862745098,0.466666666666667,0.705882352941177} 
\definecolor{colorgreen}{rgb}{0.172549019607843,0.627450980392157,0.172549019607843} 
\definecolor{colorred}{rgb}{0.83921568627451,0.152941176470588,0.156862745098039} 
\definecolor{colorblack}{rgb}{0,0,0} 
\definecolor{colororange}{rgb}{1,0.56,0} 
\usepgfplotslibrary{fillbetween}
\usepgfplotslibrary{colormaps}
\pgfplotsset{compat=1.16}

\pgfplotscreateplotcyclelist{matplotlib}{
  {matplotlib0},
  {matplotlib1},
  {matplotlib2},
  {matplotlib3},
  {matplotlib4},
  {matplotlib5},
  {matplotlib6},
  {matplotlib7},
  {matplotlib8},
  {matplotlib9}
}

\pgfplotsset{every axis/.append style={
    cycle list name=matplotlib,
}}

\usepackage{listings}

\definecolor{code_default}{HTML}{000000}
\definecolor{code_keyword}{HTML}{AC4142}
\definecolor{code_identifier}{HTML}{D28445}

\lstdefinelanguage{RISCV}{
  sensitive=false,
  morecomment=[l]{//},
  alsoletter={.},
  morekeywords=[1]{
    lp.setup, mv, lw, p.lw, sw, p.sw, pv.sdotsp.b, pv.shuffle2.b, p.subNR, p.addNR
  },
  morekeywords=[2]{
    zero, ra, sp, gp, tp, t0, t1, t2, t3, t4, t5, t6, s0, s1, a0, a1, a2, a3, a4, a5, a6, a7, a8, a9, a10, a11,
  },
  morestring=[b]",
  morestring=[b]',
}[strings, comments, keywords]

\lstdefinestyle{RISCV_STYLE}{
  language=RISCV,
  numbers=none,
  basicstyle=\scriptsize\ttfamily\color{code_default},
  keywordstyle=[1]\color{matplotlib0},
  keywordstyle=[2]\color{matplotlib1},
  float,
  captionpos=b,
  belowskip=-0.5cm
}

\usepackage{float}

%% file: acronym.tex
\newacronym{simd}{SIMD}{Single Instruction, Multiple Data}
\newacronym{elu}{ELU}{Exponential Linear Unit}
\newacronym{relu}{ReLU}{Rectified Linear Unit}
\newacronym{gelu}{GELU}{Gaussian Error Linear Unit}
\newacronym{rpr}{RPR}{Random Partition Relaxation}
\newacronym{mac}{MAC}{Multiply Accumulate}
\newacronym{dma}{DMA}{Direct Memory Access}
\newacronym{bmi}{BMI}{Brain--Machine Interface}
\newacronym{bci}{BCI}{Brain--Computer Interface}
\newacronym{smr}{SMR}{Sensory Motor Rythms}
\newacronym{eeg}{EEG}{Electroencephalography}
\newacronym{svm}{SVM}{Support Vector Machine}
\newacronym{svd}{SVD}{Singular Value Decomposition}
\newacronym{evd}{EVD}{Eigendecomposition}
\newacronym{iir}{IIR}{Infinite Impulse Response}
\newacronym{fir}{FIR}{Finite Impulse Response}
\newacronym{fc}{FC}{Fabric Controller}
\newacronym{nn}{NN}{Neural Network}
\newacronym{mrc}{MRC}{Multiscale Riemannian Classifier}
\newacronym{flop}{FLOP}{Floating Point Operation}
\newacronym{sos}{SOS}{Second-Order Section}
\newacronym{ipc}{IPC}{Instructions per Cycle}
\newacronym{tcdm}{TCDM}{Tightly Coupled Data Memory}
\newacronym{fpu}{FPU}{Floating Point Unit}
\newacronym{fma}{FMA}{Fused Multiply Add}
\newacronym{alu}{ALU}{Arithmetic Logic Unit}
\newacronym{dsp}{DSP}{Digital Signal Processing}
\newacronym{gpu}{GPU}{Graphics Processing Unit}
\newacronym{soc}{SoC}{System-on-Chip}
\newacronym{mi}{MI}{Motor-Imagery}
\newacronym{csp}{CSP}{Commmon Spatial Patterns}
\newacronym{fbcsp}{FBCSP}{Filter-Bank \acrlong{csp}}
\newacronym{pulp}{PULP}{parallel ultra-low power}
\newacronym{soa}{SoA}{state-of-the-art}
\newacronym{bn}{BN}{Batch Normalization}
\newacronym{isa}{ISA}{Instruction Set Architecture}
\newacronym{ecg}{ECG}{Electrocardiogram}
\newacronym{ppg}{PPG}{Photoplethysmography}
\newacronym{mcu}{MCU}{microcontroller}
\newacronym{rnn}{RNN}{recurrent neural network}
\newacronym{cnn}{CNN}{convolutional neural network}
\newacronym{tcn}{TCN}{temporal convolutional network}
\newacronym{emu}{EMU}{epilepsy monitoring unit}
\newacronym{pd}{PD}{Parkinson's disease}
\newacronym{ad}{AD}{Alzheimer's disease}
\newacronym{tueg}{TUEG}{Temple University Hospital EEG}
\newacronym{tuh}{TUH}{Temple University Hospital}
\newacronym{tusl}{TUSL}{TUH EEG Slowing Corpus}
\newacronym{tuar}{TUAR}{TUH EEG Artifact Corpus}
\newacronym{tuab}{TUAB}{TUH Abnormal EEG Corpus}

\newacronym{fm}{FM}{Foundation Model}
\newacronym{nlp}{NLP}{natural language processing}
\newacronym{gpt}{GPT}{Generative Pre-trained Transformer}
\newacronym{ssm}{SSM}{State Space Model}
\newacronym{moe}{MoE}{Mixture of Experts}
\newacronym{bimamba}{bi-Mamba}{bidirectional Mamba}
\newacronym{lejepa}{LeJEPA}{Latent-Euclidean Joint-Embedding Predictive Architecture}
\newacronym{jepa}{JEPA}{Joint-Embedding Predictive Architecture}
\newacronym{sigreg}{SIGReg}{Sketched Isotropic Gaussian Regularization}
\newacronym{ecf}{ECF}{Empirical Characteristic Function}
\newacronym{cf}{CF}{Characteristic Function}
\newacronym{tsne}{t-SNE}{T-distributed Stochastic Neighbor Embedding}
\newacronym{flops}{FLOPS}{Floating-point Operations Per Second}
\newacronym{oom}{OOM}{Out-Of-Memory}
\newacronym{ssl}{SSL}{Self-Supervised Learning}
\newacronym{fft}{FFT}{Fast Fourier Transform}

\newacronym{lda}{LDA}{Linear Discriminant Analysis}
\newacronym{sota}{SOTA}{state-of-the-art}
\newacronym{auroc}{AUROC}{Area Under the Receiver Operating Characteristic}
\newacronym{aupr}{AUPR}{Area Under the Precision-Recall}
\newacronym{balacc}{Bal. Acc}{Balanced Accuracy}


%% file: Sections/Introduction.tex
\section{Introduction}\label{sec:intro}

Physiological foundation models (FMs) have emerged as a powerful paradigm for biosignal analysis, enabling self-supervised pretraining on large unlabeled corpora, followed by task-specific fine-tuning~\cite{Jiang2024-ag, Doner2025-by, Jin2025-ve}. However, existing FMs are largely confined to single modalities--ECG-only~\cite{Jin2025-ve}, EEG-only~\cite{Jiang2024-ag}, or PPG-only~\cite{Pillai2024-xu}--primarily because the most accessible large-scale datasets are unimodal. Since the human body operates as an integrated biological system, physiological conditions manifest across a broad spectrum of signals; leveraging multiple biosignal modalities is therefore essential for holistic health assessment. Recent work corroborates this: cross-modal transfer from EEG to ECG/PPG has proven viable~\cite{Toth2025-ng}, and multimodal pretraining can bootstrap downstream performance beyond what unimodal models achieve~\cite{Chen2025-xm}. Despite its potential, multimodal biosignal modeling faces three key challenges:

\textbf{1) Late-fusion architectures inflate model size.}
Existing multimodal approaches (e.g., ECG+EEG~\cite{Ghallab2025-hc}; ECG+PPG~\cite{pmlr-v278-liu25d}; EEG/ECG/EOG/EMG~\cite{Jiang2025-sz, Fang2024-vs}) predominantly assign a dedicated encoder per modality, aggregated through joint optimization~\cite{Fang2024-vs} or two-stage training~\cite{Jiang2025-sz}. These modality-specific backbones significantly inflate model size and memory footprint, posing a substantial barrier to deployment on resource-constrained edge devices. Efficient single-encoder fusion strategies remain rare.

\textbf{2) Robustness to missing modalities.}
At inference time, not all modalities may be available--a common scenario in clinical and wearable settings. Notable efforts include PhysioOmni~\cite{Jiang2025-sz} (decoupled tokenizer and prototype alignment) and PhysioME~\cite{Lee2025-vq} (restoration decoder), but these still rely on modality-specific encoders.

\textbf{3) Edge-device feasibility.}
Most multimodal FMs remain computationally intensive, precluding deployment on ultra-low-power microcontrollers (MCUs) for continuous wearable monitoring. To date, no work has demonstrated end-to-end quantization and MCU deployment of a multimodal physiological FM.

To address these gaps, we present \textbf{PanLUNA}, a \textit{Pan}-modal extension of LUNA~\cite{Doner2025-by} that jointly processes EEG, ECG, and PPG within a single shared encoder. The key insight is that LUNA's Channel-Unification Module--originally designed for variable EEG channel configurations--naturally generalizes to cross-modal fusion by treating each modality's channels as entries in a unified query set augmented with sensor-type embeddings, yielding a compact 5.4M-parameter model pretrained on $\sim$40,000 hours of heterogeneous biosignal data.
Our contributions are:

\begin{itemize}
    \item \textbf{Single-encoder multimodal FM:} We extend LUNA's channel-unification mechanism to fuse EEG, ECG, and PPG within one encoder, avoiding the parameter overhead of late-fusion designs. PanLUNA (5.4M parameters) matches or exceeds models up to $\sim57\times$ larger on established benchmarks: $81.21\%$ balanced accuracy on TUAB abnormal EEG detection and state-of-the-art $0.7416$ balanced accuracy on HMC sleep staging. 
    \item \textbf{Missing-modality robustness:} The single-encoder, query-fusion design inherently handles arbitrary subsets of modalities at test time without architectural modification, as demonstrated by EEG-only and ECG-only evaluations on HMC.
    \item \textbf{Quantization analysis:} We conduct a systematic study of post-training quantization (PTQ) and quantization-aware training (QAT) across five cardiac benchmarks. QAT with INT8 weights recovers $\geq96\%$ of FP32 performance, while INT2 weight quantization achieves up to 16$\times$ storage reduction with graceful degradation. 
    \item \textbf{Edge deployment:} We deploy PanLUNA on the GAP9 ultra-low-power RISC-V MCU for wearables, achieving an inference latency of 325.6\,ms and energy consumption of just 18.8\,mJ for a 10-second, 12-lead ECG window--$2.4\times$ faster and $2.4\times$ more energy-efficient than comparable unimodal FMs on the same platform.
\end{itemize}

%% file: Sections/Methods.tex
\section{Methodology}
\label{sec:methods}

\subsection{Pretraining Datasets and Preprocessing}
\label{sub:datasets}
We leverage five large publicly available datasets, totaling $\sim$40,000 hours of \gls{eeg}, \gls{ecg}, and \gls{ppg} recordings, for modality-agnostic self-supervised pretraining (Table~\ref{tab:datasets}). For \gls{eeg}, we use the \gls{tueg}~\cite{Obeid2016-to} and the Siena Scalp EEG dataset~\cite{Detti2020-pn}. For \gls{ecg}, we include MIMIC-IV~\cite{PhysioNet-mimic-iv-ecg-1.0} and CODE-15\%~\cite{Ribeiro2021-nc}. Finally, PulseDB~\cite{Wang2022-ef} provides synchronized \gls{ecg} and \gls{ppg} recordings. All signals undergo modality-specific bandpass filtering (\gls{eeg}: 0.1--75\,Hz~\cite{Jiang2024-ag}; \gls{ecg}: 0.5--120\,Hz~\cite{Habib2020-gw}; \gls{ppg}: 0.5--8\,Hz~\cite{Pillai2024-xu, Lapitan2024-pk, Ismail2021-ph}) with a 4th-order Butterworth filter, followed by notch filtering at 50 or 60\,Hz. We resample to a common frequency of 256\,Hz, and apply channel-wise z-score normalization to account for the large amplitude differences across modalities (e.g., \gls{eeg} in $\mu$V vs.\ \gls{ecg} in mV). Signals are then segmented into non-overlapping 5-second windows with a patch size of 32 samples.

\subsection{Architecture}

PanLUNA builds on LUNA~\cite{Doner2025-by}, a topology-agnostic \gls{eeg} foundation model whose Channel-Unification Module projects variable electrode configurations into a fixed latent space via cross-attention between learned queries and channel-wise features, achieving permutation invariance and linear scaling with channel count. We extend this principle from topology invariance to \textit{multimodal physiological sensing}: channels from different sensors (\gls{eeg}, \gls{ecg}, \gls{ppg}) are concatenated along the channel dimension, and the cross-attention mechanism naturally fuses information across both channels and modalities through a shared set of latent queries. This design removes the requirement for paired multimodal recordings during pretraining and enables training on large-scale unimodal corpora while maintaining a single unified encoder.

\paragraph{Input representation.}
We introduce \textit{sensor-type embeddings} via a modality-specific lookup table, added to channel features at the input stage to distinguish sensing modalities. Channel positional encodings follow modality-specific strategies. For \gls{eeg}, we retain normalized 3D electrode coordinates encoded with sinusoidal embeddings as in \cite{Doner2025-by}. For \gls{ecg}, where no standardized spatial coordinate system exists for conventional 12-lead recordings, we adopt lead-angle estimates derived from anatomical measurements on 30 body scans \cite{Chen2021-gq}, constructing a spatial encoding analogous to an EEG electrode positioning. For \gls{ppg}, recorded from a single peripheral site without meaningful spatial structure, a neutral coordinate $(0,0)$ is assigned, so that the model relies primarily on the sensor-type embedding to identify the modality.

\paragraph{Encoder.}
Signals are partitioned into short temporal segments and embedded using lightweight convolutional encoders combined with frequency features from the real-valued FFT. Patch-level features are augmented with positional encodings and sensor-type embeddings before entering the Channel--Modality Unification module, where cross-attention aggregates information across both channels and sensing modalities. The resulting latent queries are refined through self-attention, enabling cross-modal interactions without explicit late-fusion components. The unified latent sequence is then processed by a patch-wise temporal Transformer with Rotary Positional Embeddings \cite{Su2021-si} to capture long-range temporal dependencies.
\paragraph{Decoding.}
During self-supervised pretraining, a reconstruction decoder attends to the encoder outputs to recover masked signal patches in a channel-specific manner. During downstream fine-tuning, this decoder is discarded and replaced by a lightweight aggregation query that produces a pooled representation, which is fed to a classification head.

\subsection{Evaluation Datasets}
We evaluate PanLUNA on established unimodal \gls{eeg}~\cite{Doner2025-by, Tegon2025-ry} and \gls{ecg}~\cite{Jin2025-ve, Liu2024-mz, Wang2025-ja} benchmarks for direct comparison with prior foundation models, and on a multimodal sleep staging task to assess cross-modal integration. Preprocessing follows Sect.~\ref{sub:datasets}, except for windowing, where task-specific protocols ensure fair comparison (e.g., 30-seconds epochs for sleep stage classification \cite{Berry2017-gk}). Table~\ref{tab:datasets} summarizes all downstream datasets.

\begin{table}[t]
    \centering
    \scriptsize
    \setlength{\tabcolsep}{2pt}
    \caption{Summary of pretraining and downstream datasets. Ch: number of channels. FS: sampling frequency. T: window length.}
    \label{tab:datasets}
    \resizebox{\columnwidth}{!}{
    \begin{tabular}{lccccc}
    \toprule
    \textbf{Dataset} & \textbf{Modality} & \textbf{\# Subjects} & \textbf{Ch.} & \textbf{FS (Hz)} & \textbf{T (s)} \\
    \midrule
    \multicolumn{6}{l}{\textit{Pretraining (self-supervised)}} \\
    TUEG~\cite{Obeid2016-to} & EEG & 14,987 & 20/22 & 250 & 5 \\
    Siena~\cite{Detti2020-pn, Detti2020-tb, PhysioNet} & EEG & 14 & 29 & 512 & 5 \\
    MIMIC-IV~\cite{PhysioNet-mimic-iv-ecg-1.0, PhysioNet} & ECG & 161,352 & 12 & 500 & 5 \\
    CODE-15\%~\cite{Ribeiro2021-nc, Ribeiro2019-fq} & ECG & 233,700 & 12 & 400 & 5 \\
    PulseDB~\cite{Wang2022-ef} & ECG,PPG & 5,361 & 2 & 125 & 5 \\
    \midrule
    \multicolumn{6}{l}{\textit{Downstream (fine-tuning)}} \\
    TUAB~\cite{Obeid2016-to} & EEG & 2,329 & 22 & 250 & 5 \\
    PTBXL-Super~\cite{PhysioNet-ptb-xl-1.0.3} & ECG & 18,617 & 12 & 500 & 10 \\
    PTBXL-Sub & ECG & 18,617 & 12 & 500 & 10 \\
    PTBXL-Form & ECG & 7,849 & 12 & 500 & 10 \\
    PTBXL-Rhythm & ECG & 18,269 & 12 & 500 & 10 \\
    CSN~\cite{PhysioNet-ecg-arrhythmia-1.0.0} & ECG & 23,026 & 12 & 500 & 10 \\
    HMC~\cite{Alvarez-Estevez2021-gz} & EEG,ECG & 151 & 5 & 256 & 30 \\
    \bottomrule
    \end{tabular}}
    \vspace{-0.0cm}
\end{table}

For EEG, we use TUAB~\cite{Obeid2016-to} (binary normal/abnormal; a subset of TUEG with no pretraining overlap). For ECG, we evaluate on four PTB-XL sub-tasks~\cite{PhysioNet-ptb-xl-1.0.3, Wagner2020-rl, PhysioNet} and CSN~\cite{PhysioNet-ecg-arrhythmia-1.0.0, Zheng2020-vm, PhysioNet} following the MERL protocol~\cite{Liu2024-mz}. For multimodal evaluation, HMC~\cite{Alvarez-Estevez2021-gz, PhysioNet-hmc-sleep-staging-1.1, PhysioNet} five-class sleep staging (splits as in~\cite{Jiang2025-sz}), using concurrent EEG and ECG channels to assess both cross-modal fusion and missing-modality robustness.

\subsection{Evaluation Setups}
We consider three adaptation strategies. \textbf{Full fine-tuning} (FF) updates all 5.4M model parameters. The \textbf{frozen encoder} (FE) setting keeps the pretrained backbone fixed, training only the classification head ($\sim$400k parameters).  \textbf{Low-Rank Adaptation (LoRA)} \cite{Hu2021-mk} injects trainable low-rank matrices (rank 16, $\sim$180k parameters, totaling $\sim$580k parameters) into selected Transformer layers while freezing the backbone.

\subsection{Quantization}
We compress fine-tuned models to INT8/INT4/INT2 using Brevitas~\cite{brevitas}, evaluating (weight, activation) configurations: (INT8,\,INT8), (INT4,\,INT4), (INT4,\,INT8), and (INT2,\,INT8) under two strategies:
\textbf{Post-Training Quantization (PTQ)} quantizes the FP32 model directly using calibration-set activation statistics.
\textbf{Quantization-Aware Training (QAT)} simulates quantization via fake-quantization~\cite{jacob2018quantization} during 15 epochs of fine-tuning, recovering accuracy lost under PTQ.

%% file: Sections/Results.tex
\section{Results}\label{sec:results}
\subsection{Query-Level Attention Visualization}
\begin{figure*}[t]
    \centering
    \includegraphics[width=0.7\linewidth]{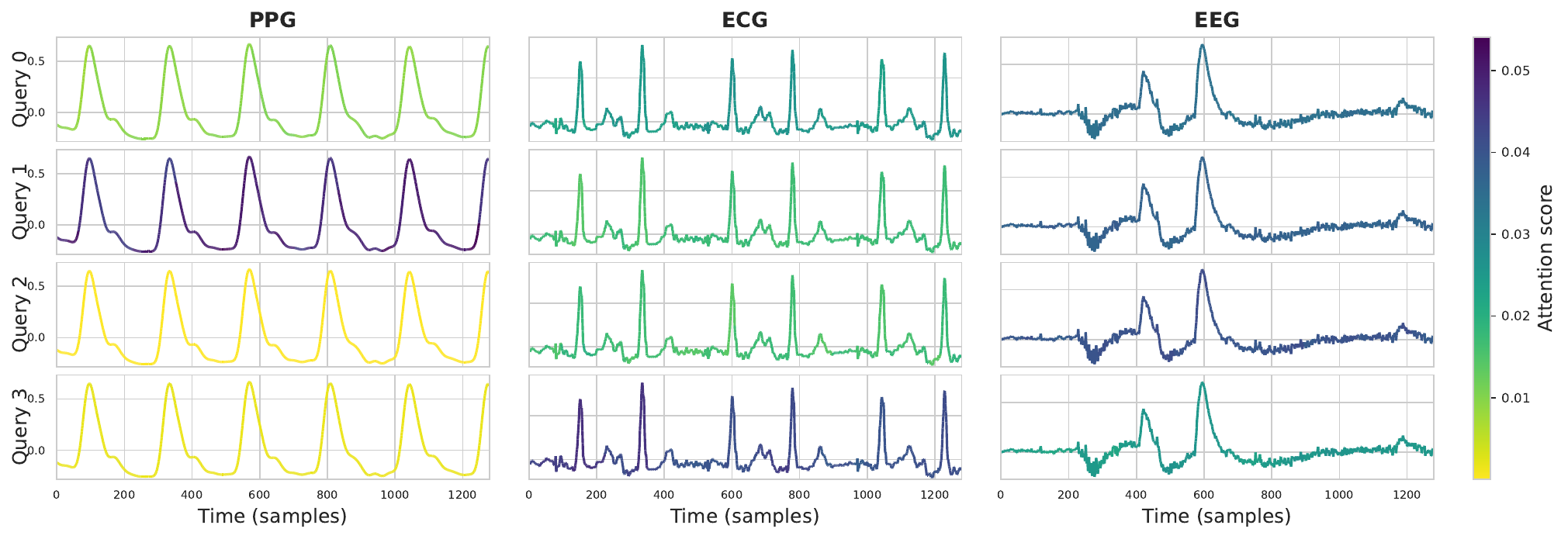}
    \vspace{-0.2cm}
    \caption{Per-patch channel-query cross-attention scores overlaid on raw signals. Each query specializes to a distinct modality: Query~3 activates on ECG QRS complexes, Query~1 on PPG waveforms, and Query~2 on EEG.}
    \label{fig:signals}
    \vspace{-0.4cm}
\end{figure*}

To examine whether the learnable queries develop modality-specific temporal sensitivity, we concatenate channels from all three modalities and visualize per-patch attention scores in Figure \ref{fig:signals} (averaged across channels per modality). The results reveal clear functional specialization: Query~3 peaks sharply at ECG QRS complexes, Query~1 dominates PPG with uniformly high attention, and Query~2 is most active on EEG. This consistent one-to-one query--modality correspondence suggests that the learnable queries specialize based on the sensor-type information encoded in the patches.

\subsection{Fine-tuning Experiments}

Table \ref{tab:tuab_results} summarizes TUAB results, where PanLUNA exceeds or achieves comparable performance to existing models. \\
Notably, under full fine-tuning, it outperforms original the EEG-only LUNA-Base and LUNA-Large models, while remaining only marginally below LUNA-Huge, despite being approximately 8 to $57\times$ smaller in parameter count. Given the architectural similarities, it is reasonable to hypothesize that PanLUNA benefits from multimodal pretraining, which likely enhances the quality of its learned representations.
\begin{table}[h!]
\scriptsize
\centering
\caption{Performance comparison on TUAB abnormal EEG detection. Best results are \textbf{bolded}.}
\label{tab:tuab_results}
\resizebox{\columnwidth}{!}{
\begin{tabular}{lllll}
\toprule
\textbf{Model} & \textbf{Size} & \textbf{Bal. Acc. (\%)$\uparrow$} & \textbf{AUC-PR$\uparrow$} & \textbf{AUROC$\uparrow$} \\
\midrule
\multicolumn{5}{l}{\textit{Self-supervised Models}} \\
BIOT \cite{Yang2023-tw}          & 3.2M   & 79.59 $\pm$ 0.57 & 0.8692 $\pm$ 0.0023   & 0.8815 $\pm$ 0.0043 \\
FEMBA-Huge \cite{Tegon2025-ry}   & 386M   & 81.82 $\pm$ 0.16 & 0.9005 $\pm$ 0.0017   & 0.8921 $\pm$ 0.0042 \\
CEReBrO-Huge \cite{Dimofte2025-gw}      & 85.15M & 81.67 $\pm$ 0.23 & 0.9049 $\pm$ 0.0026   & 0.8916 $\pm$ 0.0038 \\
LaBraM-Base \cite{Jiang2024-ag}  & 5.9M   & 81.40 $\pm$ 0.19 & 0.8965 $\pm$ 0.0016   & 0.9022 $\pm$ 0.0009 \\
LaBraM-Huge \cite{Jiang2024-ag}  & 369.8M & \textbf{82.58 $\pm$ 0.11} & 0.9204 $\pm$ 0.0011 & \textbf{0.9162 $\pm$ 0.0016} \\
CBraMod \cite{Wang2025-qt}      & 69.3M  & 82.49 $\pm$ 0.25 & \textbf{0.9221 $\pm$ 0.0015} & 0.9156 $\pm$ 0.0017 \\
LUNA-Base \cite{Doner2025-by} & 7M     & 80.63 $\pm$ 0.08 & 0.8953 $\pm$ 0.0016 & 0.8868 $\pm$ 0.0015 \\
LUNA-Large \cite{Doner2025-by} & 43M    & 80.96 $\pm$ 0.10 & 0.8986 $\pm$ 0.0005 & 0.8924 $\pm$ 0.0010 \\
LUNA-Huge \cite{Doner2025-by}  & 311.4M & 81.57 $\pm$ 0.11 & 0.9029 $\pm$ 0.0014 & 0.8957 $\pm$ 0.0011 \\
\midrule
\textbf{PanLUNA (FF)} & 5.4M & 81.21 $\pm$ 0.12 & 0.8999 $\pm$ 0.0030 & 0.8932 $\pm$ 0.0025 \\
\bottomrule
\end{tabular}
}
\vspace{-0.0cm}
\end{table}

In HMC sleep staging (Table \ref{tab:hmc_results}), PanLUNA achieves its strongest performance, surpassing EEG-only models such as LaBraM-Base and CBraMod, as well as the multimodal PhysioOmni, by a relative margin of $1.27\%$ in Balanced Accuracy, $0.95\%$ in Cohen's Kappa, and $0.31\%$ in Weighted F1-Score. Evaluating ECG-only and ECG+EEG variants further highlights the architecture's robustness to missing modalities, thanks to its single-encoder and query-fusion design.

\begin{table}[h!]
\centering
\scriptsize
\setlength{\tabcolsep}{2pt}
\caption{The results of different methods on HMC (sleep stage detection). Best results are \textbf{bolded}; second-best results are \underline{underlined}. Quantization results are obtained for INT8/INT8 combination in weights and activations.}
\label{tab:hmc_results}
\resizebox{\columnwidth}{!}{
\begin{tabular}{lccccc}
\toprule
\textbf{Method} & \textbf{Training Modality} & \textbf{Test Modality} & \textbf{Balanced Accuracy} & \textbf{Cohen's Kappa} & \textbf{Weighted F1} \\ \midrule
EEG-Conformer \cite{Song2023-lk} & EEG & EEG & $0.6767 \pm 0.0200$ & $0.5886 \pm 0.0397$ & $0.6550 \pm 0.0463$ \\
BIOT \cite{Yang2023-tw} & EEG & EEG & $0.6862 \pm 0.0041$ & $0.6295 \pm 0.0113$ & $0.7091 \pm 0.0147$ \\
LaBraM-Base \cite{Jiang2024-ag} & EEG & EEG & 0.7286 $\pm$ 0.0101 & $0.6812 \pm 0.0073$ & $0.7554 \pm 0.0024$ \\
CBraMod \cite{Wang2025-qt} & EEG & EEG & $0.7177 \pm 0.0072$ & $0.6653 \pm 0.0057$ & $0.7388 \pm 0.0052$ \\
PhysioOmni \cite{Jiang2025-sz} & EEG+EOG+EMG & EEG & \underline{0.7289 $\pm$ 0.0010} & \underline{0.6880 $\pm$ 0.0097} & \underline{0.7635 $\pm$ 0.0053} \\
\midrule
\textbf{PanLUNA (FF)} & EEG+ECG+PPG & EEG & \textbf{0.7416 $\pm$ 0.0068} & \textbf{0.6946 $\pm$ 0.0110} & \textbf{0.7659 $\pm$ 0.0086}\\
\textbf{PanLUNA (FF)} & EEG+ECG+PPG & ECG & 0.2977 $\pm$ 0.0055 & 0.1095 $\pm$ 0.0062 & 0.2876 $\pm$ 0.0092 \\ 
\textbf{PanLUNA (FF)} & EEG+ECG+PPG & EEG+ECG & 0.7002 $\pm$ 0.0217 & 0.6561 $\pm$ 0.0145 & 0.7383 $\pm$ 0.0118 \\ 
\midrule 
\multicolumn{6}{l}{\textit{Quantization Results}} \\
PanLUNA (FP32) & EEG+ECG+PPG & EEG+ECG & 0.7170 & 0.6699 & 0.7499 \\
PanLUNA (PTQ) & EEG+ECG+PPG & EEG+ECG & 0.4221 & 0.3162 & 0.4479 \\
PanLUNA (QAT) & EEG+ECG+PPG & EEG+ECG & \textbf{0.7347} & \textbf{0.6913} & 0.7273 \\
\bottomrule
\end{tabular}
}
\vspace{-0.0cm}
\end{table}

\subsection{Quantization Results}
In Table \ref{tab:quat_lora}, we report quantization results across unimodal cardiac experiments. FP32 performance is obtained using LoRA fine-tuning on a single seed, achieving SoA results on 2/5 tasks (Super and CSN) and near-best performance on Rhythm and Sub \cite{Liu2024-mz}, \cite{Jin2025-ve}, \cite{Wang2025-ja}. Using PTQ with INT8, we recover roughly $85-95\%$ of the FP32 performance without additional training. QAT with INT2 weights and INT8 activations achieves a similar fraction, enabling extremely aggressive compression: 2-bit weights reduce storage requirements by factor of 16, while maintaining large fraction of FP32 performance. The largest discrepancy occurs in the Form task, with a $2-3\%$ absolute drop under QAT with INT8 activations and INT4/INT8 weights, which we attribute to the morphological nature of its label space--subtle waveform shape features (e.g., ST-segment deviations, T-wave inversions) are inherently more sensitive to weight perturbations than the coarser rhythm or diagnostic distinctions in the other tasks. A detailed investigation of quantization sensitivity across task types is left to future work. Other tasks show at most $1.5\%$ degradation. Similar trends for INT4 and INT8 weight quantization, allow model size reductions of up to 8$\times$ with minimal performance loss. 
For multimodal HMC setting -- combining 4 EEG channels with one ECG channel (Table \ref{tab:hmc_results}) -- INT8 QAT not only preserves but marginally exceeds FP32 performance on Balanced Accuracy and Cohen's Kappa, demonstrating that aggressive compression is viable even in heterogeneous, multi-sensor configurations.
\begin{table}[h!]
\centering
\setlength{\tabcolsep}{3pt}
\caption{Quantization Results on Cardiac Experiments. \textbf{Bold} results are closest to the FP32 performance.}
\label{tab:quat_lora}
\resizebox{\columnwidth}{!}{
\begin{tabular}{l c cc cc cc cc}
\toprule
\textbf{Task} & \makecell{\textbf{AUROC} \\ \textbf{FP32 (FE)}}
& \multicolumn{2}{c}{\makecell{\textbf{W: INT8} \\ \textbf{A: INT8}}} 
& \multicolumn{2}{c}{\makecell{\textbf{W: INT4} \\ \textbf{A: INT8}}} 
& \multicolumn{2}{c}{\makecell{\textbf{W: INT2} \\ \textbf{A: INT8}}} 
& \multicolumn{2}{c}{\makecell{\textbf{W: INT4} \\ \textbf{A: INT4}}} \\

\cmidrule(lr){3-4} \cmidrule(lr){5-6} \cmidrule(lr){7-8} \cmidrule(lr){9-10}
& & \textbf{PTQ} & \textbf{QAT} 
& \textbf{PTQ} & \textbf{QAT} 
& \textbf{PTQ} & \textbf{QAT} 
& \textbf{PTQ} & \textbf{QAT} \\
\midrule

PTBXL-Super & 0.9083 
& 0.8448 & \textbf{0.9012}
& 0.5892 & 0.9011 
& 0.4976 & 0.8675
& 0.4993 & 0.8514 \\

PTBXL-Sub & 0.8880
& 0.7877 & 0.8784
& 0.6718 & \textbf{0.8797} 
& 0.4877 & 0.8272
& 0.5310 & 0.8170 \\

PTBXL-Form & 0.8331 
& 0.7215 & 0.8009
& 0.5528 & \textbf{0.8114}
& 0.4422 & 0.6710
& 0.4663 & 0.7378 \\

PTBXL-Rhythm & 0.9641
& 0.8897 & \textbf{0.9623}
& 0.6526 & 0.9508
& 0.5443 & 0.9034
& 0.5340 & 0.9099\\

CSN & 0.9505 
& 0.8743 & \textbf{0.9507}
& 0.6256 & 0.9447
& 0.5297 & 0.8821 
& 0.5384 & 0.9109\\
\bottomrule
\end{tabular}
}
\vspace{-0.0cm}
\end{table}

\subsection{Deployment}
To validate edge feasibility, we deploy PanLUNA on the GAP9 ultra-low-power RISC-V processor~\cite{greenwaves_gap_sdk} (9-core cluster at 370\,MHz, 1.5\,MB L2 SRAM). We first deploy the QAT INT8 model fine-tuned on PTB-XL Subclass (12-lead ECG, 10-second window). While this configuration processes unimodal ECG input, the deployed encoder is the full multimodal-pretrained PanLUNA backbone--to our knowledge, the first deployment of a multimodal physiological FM on an ultra-low-power MCU for wearables. Deployment uses the BioFoundation edge framework~\cite{Fasulo2026-gx, Tegon2025-ry, Doner2025-by} for automated operator tiling, double-buffered DMA, and NE16 acceleration, supplemented with custom tiled kernels for the cross-attention projections and sensor-type embedding lookup.

\begin{table}[h!]
\centering
\scriptsize
\caption{On-device biosignal inference on ultra-low-power MCUs. All GAP9 models operate at 370\,MHz.}
\label{tab:gap9}
\setlength{\tabcolsep}{2pt}
\resizebox{\columnwidth}{!}{
\begin{tabular}{lcccccccc}
\toprule
\textbf{Model} & \textbf{Platform} & \textbf{Params} & \textbf{Signal} & \textbf{Cycles} & \textbf{Latency} & \textbf{Energy} & \textbf{Power} & \textbf{MACs/cyc} \\
\midrule
\multicolumn{9}{l}{\textit{Task-specific models}} \\
AF-LSTM~\cite{Akbari2025-pm} & STM32L4 & $<$50k & ECG (1) & -- & 143\,ms & 3.5\,mJ & -- & -- \\
CLTC~\cite{Huang2024-ur} & STM32F7 & $\sim$240k & ECG (12) & -- & $\sim$1\,s & -- & 137.4\,mW & -- \\
\midrule
\multicolumn{9}{l}{\textit{Foundation models}} \\
TinyMyo~\cite{Fasulo2026-gx} & GAP9 & 3.56M & EMG & 291.3\,M & 785\,ms & 44.9\,mJ & 57.2\,mW & 6.65 \\
FEMBA~\cite{Tegon2025-ry} & GAP9 & 7.8M & EEG (22) & 629.4\,M & 1.70\,s & 75\,mJ & 44.1\,mW & 1.13 \\
\textbf{PanLUNA} & GAP9 & 5.4M & ECG (12) & 120.5\,M & \textbf{325.6\,ms} & \textbf{18.8\,mJ} & 60.2\,mW & \textbf{4.59} \\
\textbf{PanLUNA} & GAP9 & 5.4M & EEG+ECG (5) & 446.2\,M & 1.206\,s & 68.65\,mJ & 56.9\,mW & 3.23 \\
\bottomrule
\end{tabular}
}
\end{table}

Table \ref{tab:gap9} reports end-to-end on-device metrics measured on the GAP9 EVK board and compares against the published biosignal inference systems.
Compared to task-specific ARM Cortex-M classifiers ($<$250k parameters, 143\,ms--1\,s latency)~\cite{Akbari2025-pm, Huang2024-ur}, foundation models on GAP9 operate in a more computationally demanding regime: TinyMyo~\cite{Fasulo2026-gx} requires 785\,ms / 44.9\,mJ, and FEMBA~\cite{Tegon2025-ry} requires 1.70\,s / 75\,mJ per inference. PanLUNA achieves \textbf{325.6\,ms} inference latency and \textbf{18.8\,mJ} per 10-second, 12-lead ECG inference--$2.4\times$ faster and more energy-efficient than TinyMyo, $5.2\times$ faster than FEMBA--while being the only model pretrained across EEG, ECG, and PPG. Since the model operates on patch-level inputs (32 samples, i.e., 125\,ms at 256\,Hz), inference can be triggered after each new patch in a sliding-window fashion rather than waiting for a full 10-second buffer, yielding an effective streaming latency of 450.6\,ms (125\,ms acquisition $+$ 325.6\,ms compute). On a typical 300\,mAh / 3.7\,V wearable battery ($\sim$4.0\,kJ), this translates to approximately \textbf{24 days} of continuous inference-only ECG monitoring. To further validate true multimodal on-device inference, we deploy the INT8 QAT model fine-tuned on HMC sleep staging (4 EEG + 1 ECG channels, 30-second epochs--i.e., $3\times$ the temporal window of the ECG configuration). Despite processing a substantially longer input, PanLUNA completes inference in 1.206\,s at 68.65\,mJ, comfortably within the real-time budget of a 30-second epoch and on par with FEMBA, which processes only 5-second, single-modality EEG windows. On the same wearable battery, this yields approximately \textbf{20 days} of continuous multimodal sleep monitoring.